\documentclass[10pt,journal,compsoc]{IEEEtran}

\usepackage{amsmath}
\usepackage{amssymb}
\usepackage{amsthm}
\usepackage{booktabs}
\usepackage{enumitem}
\usepackage{graphicx}
\usepackage{makecell}
\usepackage{mathtools}
\usepackage{marvosym}
\usepackage{microtype}
\usepackage{multicol}
\usepackage{multirow}
\usepackage{ragged2e}
\usepackage{subcaption}
\usepackage{tabularx} 
\usepackage{adjustbox}
\usepackage[table,dvipsnames]{xcolor}
\usepackage{tikz}
\usepackage{wrapfig}
\usepackage{pifont}
\usepackage{xspace}
\usepackage{url}
\usepackage{fontawesome}
\usepackage{placeins} 
\usepackage{xcolor}
\usepackage[normalem]{ulem}
\usepackage[pagebackref=false,colorlinks,urlcolor=magenta]{hyperref}

\usepackage[capitalize]{cleveref}
\crefname{section}{Sec.}{Secs.}
\Crefname{section}{Section}{Sections}
\Crefname{table}{Table}{Tables}
\crefname{table}{Tab.}{Tabs.}
\definecolor{omnilidar}{RGB}{248,216,234}

\ifCLASSOPTIONcompsoc
  \usepackage[nocompress]{cite}
\else
  \usepackage{cite}
\fi

\begin{document}

\title{OmniLiDAR: A Unified Diffusion Framework for Multi-Domain 3D LiDAR Generation}

\author{
Youquan~Liu, 
Weidong~Yang,
Ao~Liang,
Xiang~Xu,
Lingdong~Kong,
Yang~Wu,
Dekai~Zhu,
Xin~Li,\\
Runnan~Chen,
Ben~Fei,
Tongliang~Liu,
Wanli~Ouyang
\IEEEcompsocitemizethanks{
\IEEEcompsocthanksitem Y. Liu and W. Yang are with the College of Computer Science and Artificial Intelligence, Fudan University, Shanghai, China.
\IEEEcompsocthanksitem L. Kong and A. Liang are with the School of Computing, Department of Computer Science, National University of Singapore, Singapore.
\IEEEcompsocthanksitem X. Xu is with the College of Computer Science and Technology, Nanjing University of Aeronautics and Astronautics, Nanjing, China.
\IEEEcompsocthanksitem D. Zhu is with the Technical University of Munich, Munich, Germany.
\IEEEcompsocthanksitem Y. Wu is with Nanjing University of Science and Technology, Nanjing, China.
\IEEEcompsocthanksitem X. Li is with Shanghai AI Laboratory, Shanghai, China.
\IEEEcompsocthanksitem R. Chen and T. Liu are with the University of Sydney, Sydney, Australia.
\IEEEcompsocthanksitem B. Fei and W. Ouyang are with The Chinese University of Hong Kong, Hong Kong SAR, China.

\IEEEcompsocthanksitem W. Yang, B. Fei, and R. Chen are the corresponding authors.
}
}

\IEEEtitleabstractindextext{
    \begin{abstract}
LiDAR scene generation is increasingly important for scalable simulation and synthetic data creation, especially under diverse sensing conditions that are costly to capture at scale. Typically, diffusion-based LiDAR generators are developed under \emph{single-domain} settings, requiring separate models for different datasets or sensing conditions and hindering unified, controllable synthesis under heterogeneous distribution shifts. To this end, we present \textbf{OmniLiDAR}, a unified text-conditioned diffusion framework that generates LiDAR scans in a shared range-image representation across \textbf{eight representative domains} spanning three shift types: adverse weather, sensor-configuration changes (\textit{e.g.}, reduced beams), and cross-platform acquisition (vehicle, drone, and quadruped). To enable training a single model over heterogeneous domains without isolating optimization by domain, we introduce a \textbf{Cross-Domain Training Strategy (CDTS)} that mixes domains within each mini-batch and leverages conditioning to steer generation. We further propose \textbf{Cross-Domain Feature Modeling (CDFM)}, which captures directional dependencies along azimuth and elevation axes to reflect the anisotropic scanning structure of range images, and \textbf{Domain-Adaptive Feature Scaling (DAFS)} as a lightweight modulation to account for structured domain-dependent feature shifts during denoising. In the absence of a public consolidated benchmark, we construct an \textbf{8-domain} dataset by combining real-world scans with physically based weather simulation and systematic beam reduction while following official splits. Extensive experiments demonstrate strong generation fidelity and consistent gains in downstream use cases, including generative data augmentation for LiDAR semantic segmentation and 3D object detection, as well as robustness evaluation under corruptions, with consistent benefits in limited-label regimes.
\end{abstract}

\begin{IEEEkeywords}
LiDAR scene generation, diffusion models, controllable generation, multi-domain learning, generative data augmentation.
\end{IEEEkeywords}

}

\maketitle

\IEEEdisplaynontitleabstractindextext
\IEEEpeerreviewmaketitle

\section{Introduction}
\label{sec:introduction}
\begin{figure*}[t]
    \centering
    \includegraphics[width=\textwidth]{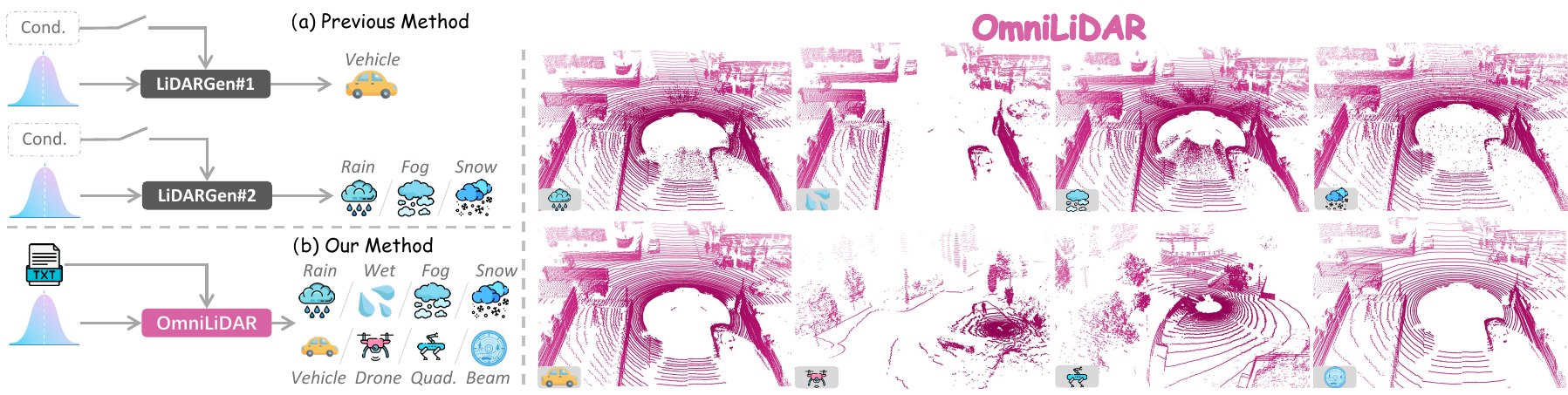}
    \caption{OmniLiDAR in context: unified multi-domain LiDAR generation with a single text-conditioned diffusion model.
    \textbf{(a)} Prior diffusion-based LiDAR generators are typically developed under \emph{single-domain} settings, \textit{i.e.}, one model per dataset or a narrowly scoped variation axis (\emph{e.g.}, weather-only), which limits coverage and scalability.
    \textbf{(b)} OmniLiDAR trains a \emph{single} generator over \textbf{eight representative domains} spanning three shift types: weather effects, sensor-configuration changes (\emph{e.g.}, beam reduction), and acquisition platforms (vehicle, drone, quadruped), and uses concise text prompts to select the target domain at inference.
    \textbf{Right:} example scans from the eight domains illustrating characteristic domain-specific patterns. }
    \label{fig:teaser}
\end{figure*}

LiDAR is a core sensing modality for autonomous driving and mobile robotics, providing accurate 3D geometry and being largely insensitive to illumination variations~\cite{liu2023uniseg,li2020deep,li2023logonet,onevl,worldlens,xie2025drivebench,wang2025monomrn}. In real-world deployment, however, LiDAR data exhibit pronounced distribution shifts caused by multiple factors, including adverse \textit{weather} (\emph{e.g.}, fog, snow, rain, and wet ground), varying \textit{sensor configurations} (\emph{e.g.}, beam count and vertical resolution), and different \textit{acquisition platforms} (\emph{e.g.}, vehicles, aerial drones, and quadruped robots). These shifts induce structured changes in return statistics, sampling geometry, and point-density patterns~\cite{kong2023robo3D,zhao2024triplemixer,hahner2021fog,hahner2022lidar,liang2025perspective,wang2025nuc-net,peng2024sam}. Since many such conditions are infrequent and costly to capture with dense annotations~\cite{bijelic2020seeing,liu2023seal,sun2020waymo}, scalable LiDAR generative modeling is increasingly attractive for simulation and synthetic data creation.

Among generative models, diffusion-based LiDAR generation methods~\cite{zyrianov2022lidargen,nakashima2024r2dm,ran2024lidm,xu2025u4d,zhu2025spiral} have achieved strong fidelity, but they are typically developed under \emph{single-domain} settings: both training and evaluation are confined to a single dataset or a narrowly scoped condition set (\emph{e.g.}, a small weather subset), as illustrated in \Cref{fig:teaser}(a). 
In practice, extending such generators to heterogeneous sensing scenarios often requires training separate models for different domains or restricting the scope of variation to maintain quality \cite{kong2023lasermix,kong2025lasermix++}. This paradigm not only scales poorly as the number of target domains grows, but also complicates model maintenance, deployment, and fair cross-domain comparison \cite{hao2024is,xie2025benchmarking,li2024optimizing,kong2025largead}. More importantly, it prevents a single controllable generator from leveraging LiDAR geometric priors and scanning regularities shared across domains. A central question therefore remains: \emph{Can a single generative model cover heterogeneous LiDAR domains induced by weather, sensor, and platform shifts, while maintaining explicit controllability over the target domain?}

Building such a unified generator is non-trivial. First, the domain gaps are inherently structured: adverse weather introduces condition-dependent attenuation and dropout patterns, reduced-beam sensors modify the angular sampling lattice, and cross-platform acquisition alters viewpoints and occlusion statistics~\cite{wu2025weathergen,liang2025perspective,kong2023robo3D,kong20253d,li20253eed,xu2025limoe}. Second, unified training must learn cross-domain shared representations while preserving domain-specific characteristics \cite{xu2025superflow++,liang2025perspective,kong2023conDA,li2024place3d,bian2025dynamiccity}. Domain-homogeneous mini-batches restrict cross-domain interaction during optimization, limiting the learning of representations shared across domains that are required for a single controllable generator.

Conversely, indiscriminate domain mixing can introduce interference across heterogeneous statistics, degrading the model’s ability to preserve domain-specific structures. A unified model must therefore learn geometry-aligned structure shared across domains, accommodate domain-dependent statistics, and enable reliable domain control within a single generative process.

To address these challenges, we propose \textbf{OmniLiDAR}, a unified, text-conditioned diffusion framework for controllable LiDAR generation across heterogeneous domains. OmniLiDAR operates on a shared \emph{range-image} representation that respects spinning-LiDAR acquisition geometry and provides a consistent modeling space across domains. Concise text descriptors serve as compact domain-level conditioning signals to specify the desired domain. This design targets descriptor-conditioned domain control rather than open-vocabulary language supervision.
We introduce a \textbf{Cross-Domain Training Strategy (CDTS)} that constructs mixed-domain mini-batches to strengthen cross-domain interaction while relying on conditioning to steer domain-specific generation. 
In addition, OmniLiDAR incorporates \textbf{Cross-Domain Feature Modeling (CDFM)} to model geometry-aligned long-range dependencies shared across domains along the azimuth and elevation axes, and \textbf{Domain-Adaptive Feature Scaling (DAFS)} as a lightweight modulation to calibrate domain-dependent statistics during denoising.

Training and evaluating a single generator across heterogeneous domains requires a unified multi-domain benchmark. However, existing public LiDAR datasets~\cite{behley2019semantickitti,bijelic2020seeing,liang2025perspective,zhao2024triplemixer,caesar2020nuscenes} each cover only a subset of the above factors and are not designed for unified multi-domain generation under a consistent protocol.

To bridge this gap, we construct an \textbf{eight-domain} LiDAR dataset that spans \textbf{three domain shift types}: 
\begin{itemize}
    \item Cross-platform clean domains (vehicle, drone, and quadruped) from public benchmarks~\cite{behley2019semantickitti,liang2025perspective}.

    \item Controlled weather-induced degradations generated via physically based simulation~\cite{kong2023robo3D,kilic2025lidar}.

    \item Systematic beam reduction implemented by deterministic subsampling of vertical scan lines.
\end{itemize}
The resulting benchmark covers \emph{eight representative domains} rather than the full Cartesian product of all factors, and focuses on unified generation under representative weather, sensor-configuration, and platform shifts within a common spinning-LiDAR setting. It strictly follows the official data splits of the source datasets.

We validate OmniLiDAR through comprehensive experiments on both generation quality and downstream utility, including LiDAR semantic segmentation, robustness evaluation under corruption shifts, cross-platform 3D object detection, and reduced-beam perception, with a focus on limited-label regimes.
Our main contributions are:
\begin{itemize}
    \item We present \textbf{OmniLiDAR}, a unified text-conditioned diffusion framework for controllable LiDAR generation across eight representative domains spanning weather, sensor-configuration, and platform shifts.
    
    \item We introduce \textbf{CDTS} for effective unified diffusion training over heterogeneous domains, together with \textbf{CDFM} for geometry-aligned directional feature modeling and \textbf{DAFS} for domain-adaptive statistical calibration within a single generator.
    
    \item We construct an \textbf{eight-domain} LiDAR dataset by combining real scans and physically based simulations (weather effects and beam reduction) under a consistent protocol and official splits.
    
    \item We demonstrate strong generation fidelity and consistent downstream benefits across multiple evaluation settings, highlighting the practical value of unified LiDAR generation, especially in limited-label regimes.
\end{itemize}
\section{Related Work}
\label{sec:related_work}

\subsection{LiDAR Scene Understanding}
LiDAR scene understanding aims to infer semantic and geometric structure from sparse and irregular 3D measurements~\cite{feng2020deep,liu2023seal,chen2023clip2scene}. Existing approaches encode structural priors through alternative data representations. Point-based methods process unordered point sets with permutation-invariant architectures~\cite{qi2017pointnet,qi2017pointnet++,hu2020randlanet,shuai2021baflac}. BEV projections map point clouds onto the ground plane to form regular spatial grids~\cite{zhang2020polarnet,zhou2021panoptic-polarnet}. Range-image representations exploit the native scanning geometry for efficient 2D processing~\cite{ando2023rangevit,kong2023rangeformer,li2025seeground,xu2025frnet,xu2020squeezesegv3,milioto2019rangenet++}. Voxel-based methods discretize 3D space into sparse grids suitable for submanifold convolutions~\cite{choy2019minkunet,zhu2021cylinder3d,yan2018second,hong2021dsnet,yin2021centerpoint}. Hybrid frameworks integrate multiple representations to combine complementary cues~\cite{tang2020spvcnn,xu2021rpvnet,liong2020amvnet,zhuang2021pmf,liu2023uniseg}. Among these representations, range images provide a structured domain aligned with spinning-LiDAR acquisition, making them a natural choice for diffusion-based generation and for modeling anisotropic dependencies along azimuth and elevation, as adopted in this work.

\subsection{LiDAR Scene Generation}
Diffusion models~\cite{latentdiffusion,ddpm,ho2022imagen,peebles2023scalable,wan2025wan,zhang2023adding,yang2023diffusion,esser2024scaling,gao2025longvie,fei2025getmesh,wang2026adasformer} have enabled high-fidelity visual generation and controllable synthesis, motivating efforts to extend diffusion processes to LiDAR data with sparse and irregular measurements. LiDARGen~\cite{zyrianov2022lidargen} introduces diffusion-based LiDAR generation built on range-image representations. UltraLiDAR~\cite{xiong2023ultralidar} adopts a voxelized VQ-VAE to learn compact latents for LiDAR synthesis and completion. R2DM~\cite{nakashima2024r2dm} formulates diffusion on range images and highlights the role of spatial inductive biases in preserving geometric fidelity. LiDM~\cite{ran2024lidm} incorporates geometry-aware constraints to better maintain surface continuity, while RangeLDM~\cite{hu2024rangeldm} employs latent diffusion for more efficient synthesis. Recent works further explore different conditioning signals for controllable generation. Text2LiDAR~\cite{wu2024text2lidar} studies text-conditioned LiDAR generation. Veila~\cite{liu2025veila} generates panoramic LiDAR from a single RGB image. LaLaLiDAR~\cite{liu2025lalalidar} conditions synthesis on scene graphs, and LiDARCrafter~\cite{liang2025lidarcrafter} extends layout conditioning to temporally coherent LiDAR sequences via a layout-to-sequence diffusion formulation. LiDAR4D~\cite{lidar4d} studies LiDAR reconstruction and dynamic novel space-time view synthesis. GS-LiDAR~\cite{gslidar} studies scene-specific LiDAR simulation and rendering using Gaussian splatting. Despite this progress, prior methods are typically developed either for a single dataset or variation axis, or for scene-specific reconstruction settings. As a result, they do not directly address unified controllable generation across heterogeneous weather, sensor, and platform shifts. In contrast, we target a \emph{single} text-conditioned diffusion model trained over \textbf{eight representative domains} that span weather effects, sensor-configuration changes, and acquisition platforms (vehicle, drone, and quadruped) under a unified protocol.


\subsection{Corruption Synthesis for Robust Perception}
Safety-critical perception systems require robustness to adverse conditions, yet large-scale driving datasets~\cite{pan2020semanticposs,sun2020waymo,caesar2020nuscenes,behley2019semantickitti,unal2022scribblekitti,geiger2012kitti} are dominated by clean data. This has motivated synthetic corruption modeling for evaluation and training. In the LiDAR domain, FSRL~\cite{hahner2021fog} proposes an optics-based fog simulation for robust 3D detection, and LSS~\cite{hahner2022lidar} extends this methodology to snowfall degradation. Robo3D~\cite{kong2023robo3D} further provides a comprehensive LiDAR corruption benchmark covering weather-induced effects, external disturbances, and internal sensor failures.
Beyond physics-based simulation, image translation methods such as CycleGAN~\cite{zhu2017unpaired} and subsequent variants~\cite{liu2017unsupervised,brooks2023instructpix2pix,parmar2024one} have been widely used to synthesize adverse-weather appearances in RGB data. Recent work also leverages diffusion models for degradation synthesis. WeatherGen~\cite{wu2025weathergen} targets adverse-weather LiDAR generation, while DriveGen~\cite{lin2025drivegen} and DriveFlow~\cite{lin2025driveflow} generate degraded driving scenes using text-to-image diffusion and rectified-flow adaptation, respectively.
While prior work mainly employs synthetic corruptions for robustness benchmarking or task-specific data augmentation, we use physically based simulation to construct controlled LiDAR domains that complement real scans. This design enables unified multi-domain training and evaluation of a single generative model.

\subsection{Multi-Domain Learning}
Recent work in autonomous driving has explored multi-domain learning~\cite{lambert2020mseg,chen2023scaledet,wu2024towards,zhang2023uni3d,zhao2020object,liu2024m3net,soum2023mdt3d}, where a single model is trained jointly on data from multiple sources to improve scalability and generalization. In the LiDAR domain, M3Net~\cite{liu2024m3net} investigates universal LiDAR semantic segmentation across datasets via multi-space alignment, PPT~\cite{wu2024towards} mitigates label taxonomy discrepancies through point-based prompting, and Uni3D~\cite{zhang2023uni3d} provides a unified baseline for multi-dataset 3D object detection. These efforts investigate multi-domain learning from the perspective of discriminative LiDAR perception. By contrast, multi-domain learning for LiDAR synthesis has received considerably less attention. This gap motivates the unified multi-domain diffusion framework proposed in this work.

\section{Methodology}
\label{sec:methodology}
\begin{figure*}[t]
    \centering
    \includegraphics[width=\textwidth]{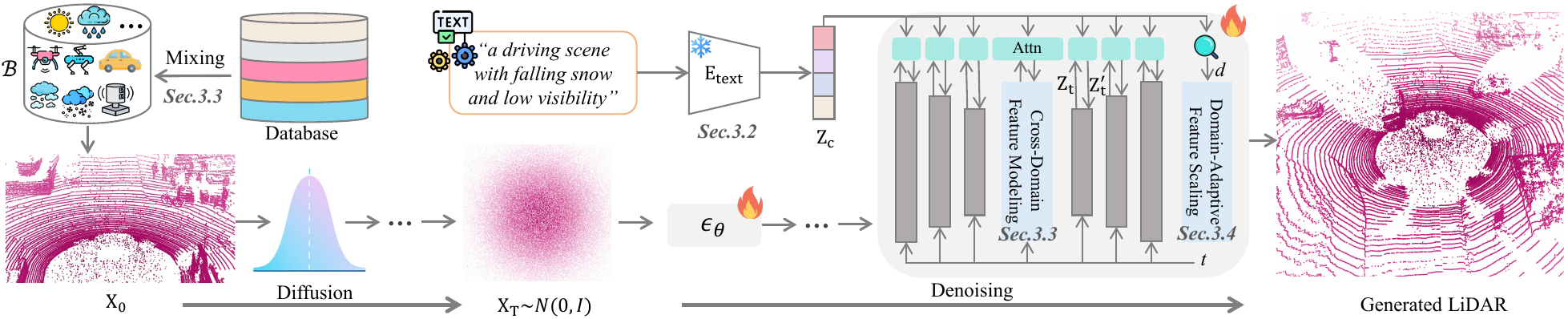}
    \vspace{-0.5cm}
    \caption{Overview of OmniLiDAR. We adopt a text-conditioned diffusion model for controllable LiDAR generation over multi-domains.
    During training, the proposed Cross-Domain Training Strategy (CDTS) mixes domains within each mini-batch, and text prompts act as domain descriptors that steer denoising via cross-attention.
    The denoiser integrates Cross-Domain Feature Modeling (CDFM) to capture geometry-aligned long-range dependencies in range images and Domain-Adaptive Feature Scaling (DAFS) to parameterize structured domain-dependent feature shifts with lightweight modulation.}
    \label{fig:framework}
\end{figure*}

We present \textbf{OmniLiDAR}, a text-conditioned diffusion framework for controllable LiDAR scene generation across heterogeneous domains. As illustrated in \Cref{fig:framework}, OmniLiDAR models LiDAR scans in a shared range-image space and uses text descriptors as domain-level conditioning signals. This enables a single generator to handle heterogeneous LiDAR distributions. We next describe the text-conditioning mechanism, along with the training strategy and architectural designs that stabilize multi-domain generation.

\subsection{Preliminary}
\label{sec:preliminary}

\noindent\textbf{LiDAR Representation.}
We represent each LiDAR scan as a point cloud $\mathcal{P}=\{\mathbf{p}^i, \mathbf{e}^i \mid i=1,\dots,N\}$, where each point has 3D coordinates $\mathbf{p}^i=(\mathbf{p}_x^i,\mathbf{p}_y^i,\mathbf{p}_z^i)\in\mathbb{R}^3$ and per-point attributes $\mathbf{e}^i\in\mathbb{R}^L$ (\emph{e.g.}, intensity and elongation). Raw point clouds are sparse and irregular, and their sampling density and return statistics vary substantially across domains due to weather effects, sensor configurations, and acquisition platforms. This heterogeneity makes it challenging to directly apply a single generator on unordered points, motivating a representation that is (i) geometrically consistent across domains, (ii) computationally efficient, and (iii) aligned with spinning-LiDAR acquisition.

Point-based representations are sensitive to domain-dependent sampling density, while voxel or occupancy grids incur cubic memory cost and tend to discard the ray-based angular structure of spinning LiDARs. Following prior works~\cite{zyrianov2022lidargen,ran2024lidm,wu2024text2lidar}, we project $\mathcal{P}$ into a range image $\mathbf{X}_0 \in \mathbb{R}^{H \times W \times 2}$ via spherical projection $\Pi:\mathbb{R}^3\mapsto\mathbb{R}^2$:
\begin{equation}
    \left( \begin{array}{c} u \\ v \end{array} \right) =
    \left( \begin{array}{c}
        \frac{1}{2}\left[1-\frac{\arctan2(\mathbf{p}_y^i,\mathbf{p}_x^i)}{\pi}\right] W\\
        \left[1-\frac{\arcsin(\mathbf{p}_z^i/r)+f_{\text{up}}}{f}\right] H
    \end{array} \right),
\end{equation}
where $(u,v)$ are pixel coordinates, $H$ and $W$ are vertical and horizontal resolutions, $r=\|\mathbf{p}^i\|_2$ is the range, and $f=|f_{\text{up}}|+|f_{\text{down}}|$ is the vertical field of view. Each pixel in $\mathbf{X}_0$ stores the range and intensity of the nearest return.

\noindent\textbf{Conditional Diffusion Models.}  
Diffusion probabilistic models~\cite{ddpm,latentdiffusion} learn a denoising model $\epsilon_\theta$ that estimates the Gaussian noise injected into the data during a forward noising process. Given a conditioning signal $c$ (\emph{e.g.}, a class label, image, or text embedding), the model is trained to predict the noise at each diffusion timestep $t \in \{1,\dots,T\}$. The standard training objective minimizes the mean-squared error between the predicted noise $\epsilon_\theta(\mathbf{X}_t, t, c)$ and the ground-truth noise $\epsilon$:
\begin{equation}
    \mathcal{L}_{\text{diff}} =
    \mathbb{E}_{t,\, \mathbf{X}_0,\, \epsilon \sim \mathcal{N}(0,I)}
    \left[
        \big\|\epsilon - \epsilon_\theta(\mathbf{X}_t, t, c)\big\|_2^2
    \right],
    \label{eq:diffusion_loss}
\end{equation}
where $\mathbf{X}_0$ denotes the clean range image and $\mathbf{X}_t$ its noisy counterpart at timestep $t$. The conditioning signal $c$ steers the denoising process toward the desired target domain, enabling controllable generation under diverse conditions.

\subsection{Text-Conditioned LiDAR Generator}
\label{sec:text_cond}
To enable controllable LiDAR synthesis across heterogeneous domains, the generator requires an explicit mechanism to specify the target domain (one of the predefined domains in our suite, spanning weather effects, sensor-configuration changes, or acquisition platforms). These domain shifts affect sampling geometry, sparsity patterns, and visibility statistics, and cannot be reliably inferred from the noisy state $\mathbf{X}_t$ alone.

We associate each domain with a short textual descriptor (\emph{e.g.}, ``a driving scene with falling snow and low visibility'', ``an outdoor scene from a drone viewpoint''). We do not target open-vocabulary or compositional semantic conditioning. These descriptors are not per-scan natural-language captions, but serve as compact domain descriptors for specifying the target domain. Each domain is associated with a small prompt pool. During training, one prompt is randomly sampled from the corresponding pool, while during inference, the first prompt is used. Each descriptor is encoded by a frozen CLIP~\cite{radford2021clip} text encoder $\text{E}_{\text{text}}$, yielding token embeddings $\mathbf{Z}_c \in \mathbb{R}^{B \times L_c \times d}$. Domain conditioning is injected through cross-attention (Attn) layers in the UNet denoiser $\epsilon_\theta$, where text embeddings interact with intermediate feature maps $\mathbf{Z}_t \in \mathbb{R}^{B \times H_t \times W_t \times C}$ across multiple resolutions. Concretely, we reshape the feature map to a sequence $\tilde{\mathbf{Z}}_t \in \mathbb{R}^{B \times L_t \times C}$ with $L_t = H_t W_t$, and compute:
\begin{equation}
\label{eq:cross_attn}
    \mathbf{Q} = \tilde{\mathbf{Z}}_t \mathbf{W}_Q,\quad
    \mathbf{K} = \mathbf{Z}_c \mathbf{W}_K,\quad
    \mathbf{V} = \mathbf{Z}_c \mathbf{W}_V,
\end{equation}
where $\mathbf{W}_Q, \mathbf{W}_K, \mathbf{W}_V$ are learnable projections. The cross-attention update is:
\begin{equation}
    \tilde{\mathbf{Z}}_t^{\,\prime}
    =
    \mathrm{Attn}(\mathbf{Q}, \mathbf{K}, \mathbf{V}) =\mathrm{softmax}\!\big(\tfrac{\mathbf{Q}\mathbf{K}^\top}{\sqrt{d_k}}\big)\,\mathbf{V},
\end{equation}
where $d_k$ denotes the query/key dimension, followed by a residual connection and feed-forward refinement. The updated sequence $\tilde{\mathbf{Z}}_t^{\,\prime}$ is reshaped back to $\mathbf{Z}_t^{\,\prime}\in\mathbb{R}^{B \times H_t \times W_t \times C}$. By inserting cross-attention blocks at multiple resolutions, the denoising process is steered toward the domain specified by the descriptor, enabling conditioning-driven domain control within a single generator.

\subsection{Cross-Domain Training Strategy}
\label{sec:cross_domain_training}

Training a unified generative model across heterogeneous LiDAR domains is challenging because domains exhibit distinct depth statistics, sparsity patterns, and visibility characteristics. These discrepancies induce shifts in feature statistics and gradient directions, making optimization sensitive to mini-batch composition.

Existing multi-dataset training schemes for autonomous driving perception often adopt \emph{domain-homogeneous} batching~\cite{liu2024m3net,zhang2023uni3d,wu2024towards}, where each mini-batch is drawn from a single domain. While this reduces intra-batch variation, it can be suboptimal for unified generative modeling: the denoiser is optimized toward one domain at a time, leading to alternating-domain updates and large step-wise distribution shifts (\emph{e.g.}, depth ranges, beam patterns, and weather-induced degradations). Such dynamics can destabilize training and limit effective cross-domain sharing.

We instead employ \emph{mixed-domain} mini-batches by sampling from the pooled training set. Let $\{\mathcal{D}_k\}_{k=1}^{M}$ denote $M$ domains, and let $\mathcal{D}=\bigcup_{k=1}^{M}\mathcal{D}_k$ denote the union of all training samples paired with their corresponding textual conditions. A mini-batch $\mathcal{B}$ is constructed as:
\begin{equation}
\mathcal{B}=\{(\mathbf{X}_0^{(i)},c^{(i)})\}_{i=1}^{B},\qquad
(\mathbf{X}_0^{(i)},c^{(i)})\sim \mathcal{D}.
\end{equation}
In our implementation, mini-batches are sampled from the concatenated multi-domain training set without explicit domain reweighting (\textit{i.e.}, following the empirical distribution of the concatenated set). Mixed-domain batching provides three benefits: (i) each update aggregates gradients from multiple domains, improving optimization stability; (ii) it reduces step-wise distribution shifts caused by alternating domain-homogeneous updates; and (iii) it promotes conditioning-driven domain control by reducing unintended entanglement of domain cues in shared parameters. The training objective remains the standard conditional diffusion loss in Eq.~\eqref{eq:diffusion_loss}.

\subsection{Cross-Domain Feature Modeling with Mamba}
\label{sec:mamba}

\begin{figure}[t]
    \centering
    \includegraphics[width=0.98\linewidth]{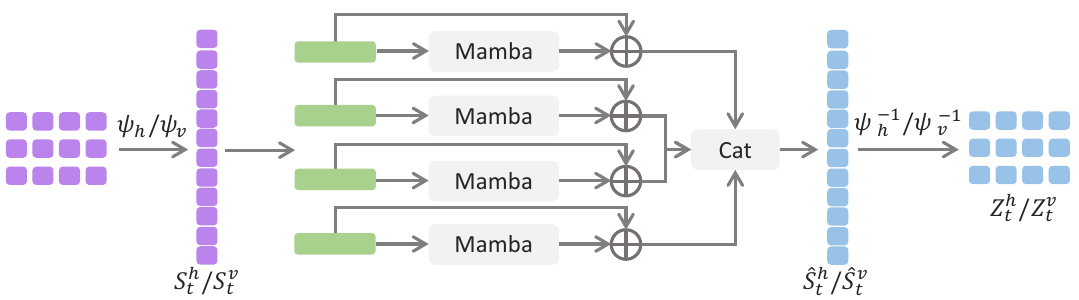}
    \vspace{-0.3cm}
    \caption{Cross-Domain Feature Modeling with directional sequence modeling. }
    \vspace{-0.3cm}
    \label{fig:mamba}
\end{figure}

Mixed-domain training improves optimization stability, but the denoiser must also capture long-range geometric structure in the shared range-image space. The key challenge is the anisotropic structure of LiDAR range images: along azimuth, measurements follow a continuous sweep and exhibit long-range correlations, while along elevation, they correspond to discrete scan rings with structured discontinuities. Purely local convolutions have limited capacity to model such dependencies, whereas global self-attention is often computationally expensive at high resolution and does not directly incorporate scan-aligned inductive biases. We therefore introduce \textbf{Cross-Domain Feature Modeling (CDFM)}, which uses scan-aligned restructuring and Mamba-based directional sequence modeling along the two physically meaningful scan axes. Under mixed-domain training, this design helps preserve geometry-aligned long-range dependencies that are shared across heterogeneous LiDAR domains, as illustrated in \Cref{fig:mamba}.

Given a text-conditioned feature map $\mathbf{Z}_t' \in \mathbb{R}^{B \times H_t \times W_t \times C}$ in the shared range-image space, CDFM explicitly constructs two sequences aligned with the LiDAR scanning geometry:
\begin{equation}
    \mathbf{S}_t^{h} = \psi_h(\mathbf{Z}_t') \in \mathbb{R}^{B \times L \times C},
    \qquad
    \mathbf{S}_t^{v} = \psi_v(\mathbf{Z}_t') \in \mathbb{R}^{B \times L \times C},
\end{equation}
where $L = H_t W_t$. The operators $\psi_h$ and $\psi_v$ convert the 2D feature map into scan-aligned sequences $\mathbf{S}_t^{h}$ and $\mathbf{S}_t^{v}$ by flattening along azimuth and elevation, respectively. Concretely,
\begin{equation}
\begin{aligned}
\psi_h(\mathbf{Z}_t')[b,\, vW_t+u,\, :] &= \mathbf{Z}_t'[b,\, v,\, u,\, :],\\
\psi_v(\mathbf{Z}_t')[b,\, uH_t+v,\, :] &= \mathbf{Z}_t'[b,\, v,\, u,\, :],
\end{aligned}
\end{equation}
where $b$ indexes the batch dimension, $u=0,\dots,W_t-1$, and $v=0,\dots,H_t-1$. We reuse the same Mamba layer $\mathcal{M}_\theta$ for both directional sequences, \textit{i.e.}, the horizontal and vertical passes share parameters. For $\mathbf{S}\in\{\mathbf{S}_t^{h},\,\mathbf{S}_t^{v}\}$, we apply a residual Mamba update:
\begin{equation}
    \hat{\mathbf{S}}
    = \mathbf{S}
    + \mathcal{M}_\theta\big(\mathrm{LN}(\mathbf{S})\big),
    \label{eq:mamba_update}
\end{equation}
where $\mathrm{LN}(\cdot)$ denotes layer normalization over the channel dimension. In our implementation, $\mathcal{M}_\theta$ is applied in a group-wise manner by splitting the channel dimension into $G$ groups (with $G{=}4$), applying the same Mamba layer to each group, and concatenating the group outputs along the channel dimension. This factorization constrains sequence modeling to operate within channel subspaces while controlling the parameterization. The concatenated features are followed by a channel projection, which enables cross-group channel mixing.

The updated sequences are mapped back to the spatial domain,
\begin{equation}
    \mathbf{Z}_t^{h} = \psi_h^{-1}(\hat{\mathbf{S}}_t^{h}),\qquad
    \mathbf{Z}_t^{v} = \psi_v^{-1}(\hat{\mathbf{S}}_t^{v}),
\end{equation}
and fused by averaging to obtain the final feature map. This directional two-pass design propagates information along both scan axes and helps preserve geometry-consistent structures in the synthesized range images, especially under platform- and sampling-induced structural variation. We apply CDFM only at deeper UNet stages, where long-range geometric structure is most salient.

\subsection{Domain-Adaptive Feature Scaling}
\label{sec:dafs}

Text conditioning specifies the target domain at a semantic level, and CDFM captures geometry-aware long-range structure. However, neither mechanism explicitly addresses the \emph{domain-specific feature statistics} arising from LiDAR sensing physics. Even under mixed-domain training, feature distributions can remain mismatched: weather-induced attenuation alters range--intensity statistics and dropout patterns, reduced-beam configurations change vertical sampling profiles, and cross-platform viewpoints shift occlusion and depth distributions. These variations are low-level yet highly structured, and are not reliably handled by semantic conditioning (which primarily steers high-level content) or by sequence modeling (which captures dependencies but does not calibrate per-domain statistics). Moreover, normalization layers tend to impose domain-agnostic moments, further suppressing domain-specific cues. As a result, the decoder may operate on distributionally misaligned activations across domains, limiting conditioning fidelity and degrading domain-specific characteristics.

To mitigate this mismatch, we propose \textbf{Domain-Adaptive Feature Scaling}, a lightweight modulation module that calibrates decoder activations with domain-aware affine parameters. Unlike common feature modulation schemes that are driven purely by semantic embeddings or are designed to enforce domain invariance, DAFS explicitly targets structured statistical discrepancies between LiDAR domains. Importantly, we constrain the modulation magnitude to avoid overfitting to any single domain and to maintain stable multi-domain training.

Formally, let $\mathbf{F}\in\mathbb{R}^{B \times c \times H \times W}$ denote the decoder feature map to be modulated, and let $d\in\{1,\dots,M\}$ be the domain identifier for each sample. A learnable domain embedding table produces a $2c$-dimensional vector, which is then split into channel-wise gain and bias:
\begin{equation}
    [\gamma_d \,\|\, \beta_d] = \mathrm{Embed}(d), 
    \qquad
    \gamma_d, \beta_d \in \mathbb{R}^{c}.
\end{equation}
To stabilize training under heterogeneous domains, we bound the modulation parameters:
\begin{equation}
    \tilde{\gamma}_d = \tanh(\gamma_d)\,\lambda,
    \qquad
    \tilde{\beta}_d = \tanh(\beta_d)\,\lambda,
\end{equation}
where $\lambda$ is a small constant. The calibrated feature map is obtained via channel-wise affine modulation:
\begin{equation}
    \mathbf{F}^{\,\prime}
    = (1 + \tilde{\gamma}_d)\,\mathbf{F} + \tilde{\beta}_d.
\end{equation}
DAFS adds negligible overhead, yet provides effective domain-aware statistical calibration, allowing the decoder to preserve structured domain characteristics while sharing a single set of generative parameters across domains.

Overall, OmniLiDAR combines mixed-domain training for stable supervision, text conditioning for explicit domain control, CDFM for geometry-aligned long-range modeling, and DAFS for domain-specific statistical calibration. Together, these components jointly model semantics, geometry, and feature statistics, enabling controllable and high-fidelity LiDAR generation under heterogeneous domain shifts.

\section{Experiments}
\label{sec:experiments}

In this section, we conduct extensive experiments to validate the proposed OmniLiDAR framework. We first introduce the datasets (\Cref{sec:dataset}) and summarize implementation details (\Cref{sec:detail}). We then compare OmniLiDAR with state-of-the-art methods across multiple benchmarks and tasks (\Cref{sec:comparative}). Finally, we present ablation studies to quantify the contribution of each component (\Cref{sec:ablation}).

\subsection{Datasets}
\label{sec:dataset}
A unified generator that covers shifts in weather, sensor configuration, and acquisition platform requires training data that jointly reflects these factors under a single protocol. As no public benchmark provides such a consolidated setup, we thus construct an 8-domain LiDAR dataset by combining real-world scans with physically based simulation and systematic beam reduction.


\begin{table*}[t]
    \centering
    \caption{Quantitative comparison of LiDAR generation quality across eight domains. We report distribution-based metrics, including FRD, FRID, FSVD, FPVD, JSD, and MMD, on the constructed multi-domain dataset. Lower values indicate better performance. The MMD metric is reported in $10^{-4}$.}
    \vspace{-0.2cm}
    \begin{adjustbox}{max width=\textwidth}
    \begin{tabular}{
        r|r|
        cccccc|cccccc|cccccc|cccccc
    }
    \toprule
    \multirow{2}{*}{\textbf{Method}} 
    & \multirow{2}{*}{\textbf{Venue}} 
        & \multicolumn{6}{c|}{\textbf{Vehicle}} 
    & \multicolumn{6}{c|}{\textbf{Snow}} 
    & \multicolumn{6}{c|}{\textbf{Fog}}
    & \multicolumn{6}{c}{\textbf{Rain}} 
    \\
    & & FRD & FRID & FPVD & FSVD & JSD & MMD
      & FRD & FRID & FPVD & FSVD & JSD & MMD
      & FRD & FRID & FPVD & FSVD & JSD & MMD
      & FRD & FRID & FPVD & FSVD & JSD & MMD
    \\
    \midrule\midrule

    Text2LiDAR & ECCV'24 & 456.38 & 40.05  & 28.06 & 30.33 & \textbf{0.04} & 1.07 & 617.52&56.22 &60.81 & 52.06 & 0.06&3.12 &1199.70 &151.42 &65.67 &54.35 &0.08 & 5.14&1026.63 &101.74 &58.84 &50.66 &0.12 & 15.06 \\

    R2DM & ICRA'24 & 443.31&10.69 &11.27 &12.75 & \textbf{0.04} & 2.97 & 453.14 &11.25 & 28.76 & 27.02 & 0.05 &3.32 &444.22 &11.56 &20.67 &17.17 &0.04 &4.28 &418.60 &8.52 &14.75 &15.61 &\textbf{0.03} &1.12 \\

    WeatherGen & CVPR'25 & 417.16& 9.42&9.11 &11.60 & \textbf{0.04} & 1.27&427.00 & 5.56&20.05 &19.09 &\textbf{0.03} & \textbf{0.50} &403.65 &8.80 & \textbf{14.76} & \textbf{12.12} &0.05 &1.02 &406.73 &7.00 &12.56 &12.84 & 0.04 & 1.19 \\
    
    \rowcolor{gray!10}

    \rowcolor{omnilidar}
    \textbf{OmniLiDAR} & \textbf{Ours} & \textbf{410.91} & \textbf{8.37} & \textbf{7.79} & \textbf{10.06} & \textbf{0.04} & \textbf{0.99} & \textbf{424.02} & \textbf{4.76} & \textbf{15.68} & \textbf{14.94} & 0.05& 0.86 & \textbf{396.64} & \textbf{7.21} &15.10 & 13.40 & \textbf{0.03} & \textbf{0.54} & \textbf{402.73} & \textbf{5.42} & \textbf{12.02} & \textbf{12.62} & 0.04 & \textbf{0.98}\\
    
    \midrule\midrule
    
    \multirow{2}{*}{\textbf{Method}} 
    & \multirow{2}{*}{\textbf{Venue}} 
    & \multicolumn{6}{c|}{\textbf{Wet Ground}} 
      & \multicolumn{6}{c|}{\textbf{Beam-32}} 
      & \multicolumn{6}{c|}{\textbf{Drone}}
      & \multicolumn{6}{c}{\textbf{Quadruped}} 
    \\
    & & FRD & FRID & FPVD & FSVD & JSD & MMD
      & FRD & FRID & FPVD & FSVD & JSD & MMD
      & FRD & FRID & FPVD & FSVD & JSD & MMD
      & FRD & FRID & FPVD & FSVD & JSD & MMD
    \\ \midrule\midrule

    Text2LiDAR & ECCV'24 &850.32 &58.12 &124.61 &105.32 &0.12 &3.95 &746.70 &99.11 &79.19 &69.05 &0.05 &1.83 & 2251.44 &191.12 &269.78 &260.06 &0.28 &46.73 &1711.12 &134.66 &206.64 &178.95 &0.16 & 12.90 \\

    R2DM & ICRA'24 & 455.14 &6.07 & 13.42& 16.47& \textbf{0.04} & \textbf{0.94} &396.99 &5.69 &13.27 &13.19 & \textbf{0.03} &1.00 & 1795.78 & 88.82 & 37.01 & 29.77& 0.07&1.98 &425.50 & 5.81 & 30.42 &26.78 &0.08 &1.86 \\

    WeatherGen & CVPR'25 &451.63 &6.31 &11.11 &12.97 &0.06 &1.93 &362.55 &4.84 &10.12 &9.93 &0.04 &1.03 & 1781.88 &82.19 &33.24 & \textbf{25.89} & \textbf{0.06} & \textbf{1.52} & \textbf{420.71} & 4.83&28.32 &24.93 & \textbf{0.07} &1.01 \\

    \rowcolor{omnilidar}
    \textbf{OmniLiDAR} & \textbf{Ours} & \textbf{448.83} & \textbf{4.73} & \textbf{9.87} & \textbf{12.40} &0.05 &1.21 & \textbf{352.61} & \textbf{3.98} & \textbf{9.21} & \textbf{8.99} & 0.04& \textbf{0.90} & \textbf{1748.28} & \textbf{81.09} & \textbf{31.81} & 26.34& \textbf{0.06} & 1.60& 422.44& \textbf{4.38} & \textbf{27.59} & \textbf{23.01} & \textbf{0.07} & \textbf{0.87} \\
    
    \bottomrule
    \end{tabular}
    \end{adjustbox}
    \label{tab:comparative_domains}
\end{table*}

We include three clean platform domains: Vehicle scans from SemanticKITTI~\cite{behley2019semantickitti} and Drone/Quadruped scans from Pi3DET~\cite{liang2025perspective}, consisting of real-world outdoor LiDAR captured with 64-beam sensors and without simulated weather effects. Based on the SemanticKITTI vehicle split, we further construct four weather-corrupted domains: Fog, Wet Ground, and Snow using Robo3D's physically based simulation~\cite{kong2023robo3D}, where Wet Ground is applied to estimated ground-plane returns only, and Rain using the physically based LISA simulator~\cite{kilic2025lidar} with severity levels defined following TripleMixer~\cite{zhao2024triplemixer}. For all weather-corrupted domains, each scan is assigned a severity level uniformly at random while preserving the original 64-beam configuration. To model sensor-configuration shifts within the spinning-LiDAR family, we additionally construct a Beam-32 domain by uniformly discarding every other vertical beam from SemanticKITTI, approximating a 32-beam spinning LiDAR. Each domain is associated with a small set of short prompt templates that specify only the domain factor for text conditioning. The complete prompt templates are provided in the supplementary material for reproducibility.

In addition to the constructed 8-domain dataset used for unified training, several existing public datasets are adopted for evaluation and downstream tasks. KITTI-360~\cite{liao2022kitti} is used for benchmarking LiDAR generation quality following prior works. NuScenes~\cite{caesar2020nuscenes,fong2022panoptic-nuscenes}, which provides native 32-beam LiDAR scans, together with the SemanticKITTI~\cite{behley2019semantickitti} dataset and its corrupted variant SemanticKITTI-C~\cite{kong2023robo3D}, are used for LiDAR semantic segmentation and robustness evaluation. We additionally use the real-world adverse-weather SemanticSTF~\cite{semanticstf} dataset for LiDAR semantic segmentation evaluation under dense fog, light fog, rain, and snow. Furthermore, the Drone and Quadruped benchmarks of Pi3DET~\cite{liang2025perspective} are adopted for 3D object detection experiments.

\subsection{Implementation Details}
\label{sec:detail}

\textbf{Training Setup.}
OmniLiDAR is trained as a unified diffusion model on the constructed 8-domain dataset described in \Cref{sec:dataset}. All experiments are conducted on a NVIDIA H100 GPU with a batch size of 16. We use the AdamW optimizer with a learning rate of $1\times10^{-4}$ and train the model for 500k iterations. LiDAR points from different domains are projected into range images with a unified resolution of $64\times1024$. A cosine noise schedule is adopted for diffusion training, and we use 256 sampling steps at inference. Following prior work~\cite{nakashima2024r2dm}, we linearly map intensity to $[-1,1]$ and apply log scaling to ranges followed by normalization to $[-1,1]$.

\noindent\textbf{Generative Quality Evaluation.}
For multi-domain generation evaluation, we sample 2k LiDAR scans for each domain using the trained OmniLiDAR model and report generation quality metrics following prior works~\cite{ran2024lidm,nakashima2024r2dm,liu2025veila}. For comparison on KITTI-360, we replace the Vehicle domain in the training corpus with the KITTI-360 training set while keeping all other settings unchanged. Under this setting, 10k LiDAR scans are generated for quantitative comparison.

\noindent\textbf{Generative Data Augmentation for Downstream Tasks.}
We evaluate the utility of generated samples via generative data augmentation (GDA)~\cite{liu2025lalalidar,zhu2025spiral,liu2025veila} on LiDAR semantic segmentation, robustness evaluation, and 3D object detection. For each task, we first subsample a fixed proportion of real training data (\emph{e.g.}, one scan out of every 100 for the 1\% setting), and additionally generate 10k LiDAR scans with OmniLiDAR for the corresponding domain.

Pseudo labels for generated samples are obtained offline using the same trained off-the-shelf LiDAR semantic segmentation or object detection model for each task. This pseudo-label source is fixed across all compared generative augmentation methods. Real and generated samples are then mixed to train downstream models from scratch with random initialization using standard training protocols, and all evaluations are performed on the official validation sets. For robustness evaluation, OmniLiDAR-generated samples from specific corruption domains are used to augment the clean SemanticKITTI training set, and performance is measured on SemanticKITTI-C across corruption types and severity levels. For sensor-configuration analysis, the generated Beam-32 samples are used to augment training data on nuScenes for LiDAR semantic segmentation under reduced vertical resolution. Cross-platform object detection is evaluated on the Drone and Quadruped benchmarks of Pi3DET.

\noindent\textbf{Evaluation Metrics.}
Following prior works~\cite{ran2024lidm,nakashima2025fast}, we evaluate LiDAR scene generation quality using a set of complementary distribution-based metrics, including Fréchet Range Distance (FRD), Fréchet Range Image Distance (FRID), Fréchet Sparse Volume Distance (FSVD), Fréchet Point-based Volume Distance (FPVD), Jensen--Shannon Divergence (JSD), Maximum Mean Discrepancy (MMD), and Fréchet Point Distance (FPD). For LiDAR semantic segmentation, we report the Intersection-over-Union (IoU) for each semantic class and the mean IoU (mIoU) over all classes. To evaluate robustness under adverse conditions, we follow the Robo3D benchmark and report the mean Corruption Error (mCE) score. For 3D object detection, we report the recall at an IoU threshold of 0.5 using the 11-point interpolation protocol (R11@0.5), following the official evaluation protocol of Pi3DET~\cite{liang2025perspective}.

\begin{table}[t]
    \centering
    \caption{Quantitative comparison of LiDAR scene generation methods on the \textit{KITTI-360} dataset.}
    \vspace{-0.2cm}
    \begin{adjustbox}{max width=0.98\linewidth}
    \begin{tabular}{r|r|c|c|c|c}
    \toprule
    \textbf{Method} & \textbf{Venue} & 
        \textbf{FRD}$\downarrow$ & 
    \textbf{FPD}$\downarrow$  &
   \textbf{JSD}$\downarrow$ & 
   \textbf{MMD}$\downarrow$ 
    \\
    \midrule\midrule
    LiDARGAN~\cite{caccia2019deep} & IROS'2019& 3003.80 & - & - & 30.60 \\
    LiDARVAE~\cite{caccia2019deep} & IROS'2019 & 2261.50  & - & 0.16 & 10.00 \\
    ProjectedGAN~\cite{sauer2021projected} &NeurIPS'21 & 2117.20 & - & 0.09 & 3.47 \\
    LiDARGen~\cite{zyrianov2022lidargen} & ECCV'22 & 579.39 & 90.29 & 0.07 & 7.39 \\
    LiDM~\cite{ran2024lidm} & CVPR'24 & 334.55 & 34.36 & 0.05 & \textbf{1.07} 
    \\
    R2DM~\cite{nakashima2024r2dm} & ICRA'24 & 179.43 & 6.99 & \textbf{0.03} & 1.41
    \\
    Text2LiDAR~\cite{wu2024text2lidar} & ECCV'24 & 425.90 & 11.39 & 0.06 & 1.63
    \\
    WeatherGen~\cite{wu2025weathergen} & CVPR'25 & 160.20 & 6.91 & \textbf{0.03} & 1.62 
    \\
    \rowcolor{omnilidar}
    \textbf{OmniLiDAR} & \textbf{Ours} & \textbf{158.13} & \textbf{6.89} & \textbf{0.03} & 1.12
    \\
    \bottomrule
    \end{tabular}
    \end{adjustbox}
\label{tab:kitti360_gen}
\end{table}
\begin{table}[t]
    \centering
    \caption{Generative data augmentation (GDA) for LiDAR semantic segmentation on the \textit{SemanticKITTI} dataset, using synthetic samples generated by existing methods and OmniLiDAR under different ratios (1\%, 10\%, and 20\%) of real labeled training data. Results are reported in mIoU (\%).}

    \vspace{-0.2cm}
    \resizebox{0.98\linewidth}{!}{
    \begin{tabular}{r|r|ccc|ccc}
    \toprule
     \multirow{2}{*}{\textbf{Method}} & \multirow{2}{*}{\textbf{Venue}} & \multicolumn{3}{c}{\textbf{MinkUNet}} & \multicolumn{3}{c}{\textbf{SPVCNN}} 
    \\
    & &   \textbf{1\%} & \textbf{10\%} & \textbf{20\%}& \textbf{1\%} & \textbf{10\%} & \textbf{20\%}  
    \\
    \midrule\midrule
    \rowcolor{yellow!10}
     \textit{Sup.-only} 
     & - & 40.39 & 60.90 & 62.84 & 37.86 & 59.07 & 61.16 
    \\
    LiDARGen~\cite{zyrianov2022lidargen} & ECCV'22 & 36.11 & 54.73 &  60.39 & 36.44 & 55.04 &  59.71
    \\
    Text2LiDAR~\cite{wu2024text2lidar} & ECCV'24 &  40.23 & 55.00 & 58.35 & 40.55 & 53.87 & 58.34
    \\
    R2DM~\cite{nakashima2024r2dm} & ICRA'24 & 53.38 & 60.78 & 62.57 & 50.25 & 60.11 & 62.34
    \\
    WeatherGen~\cite{wu2025weathergen} & CVPR'25 &55.14 &64.21 & 66.00 & 55.67 & 64.95 & 65.98 \\
    
    \rowcolor{omnilidar}
    \textbf{OmniLiDAR} 
    &\textbf{Ours} 
    &  \textbf{59.49}
    &\textbf{65.07}
    & \textbf{66.59}
    &  \textbf{59.73}
    & \textbf{65.99}
    &  \textbf{66.36}
    \\\bottomrule
    \end{tabular}}
    \label{tab:semkitti_seg}
    \vspace{-0.3cm}
\end{table}

\begin{table}[t]
    \centering
    \caption{Generative data augmentation (GDA) results for LiDAR semantic segmentation on the real-world \textit{SemanticSTF}~\cite{semanticstf} validation set. We augment the full real labeled training set with synthetic fog, snow, and rain samples generated by OmniLiDAR, and report mIoU (\%) on each weather subset and overall performance.}
    \vspace{-0.2cm}
    \resizebox{0.98\linewidth}{!}{
    \begin{tabular}{l|l|cccc|c}
        \toprule
       \textbf{Initial} & \textbf{Backbone} & \textbf{Dense Fog} & \textbf{Light Fog} & \textbf{Rain} & \textbf{Snow} & \textbf{All} \\
        \midrule\midrule
        \rowcolor{yellow!10}
        \textit{sup.-only} & MinkUNet & 52.5 & 55.1 & 58.6 & 54.0 & 56.2 \\
        \rowcolor{omnilidar}
        OmniLiDAR & MinkUNet & \textbf{53.2} & \textbf{56.6} & \textbf{59.0} & \textbf{55.3} & \textbf{57.9} \\
        \bottomrule
    \end{tabular}}
\label{tab:semanticstf_lidarseg}
\vspace{-0.3cm}
\end{table}
\begin{table}[t]
    \centering
    \caption{Robustness evaluation of LiDAR generative methods under four out-of-distribution corruptions in the \textit{SemanticKITTI-C} dataset. The mCE score is the lower the better while mIoU scores are the higher the better. All mCE, and mIoU scores are given in percentage (\%). Avg denotes the average mIoU scores of methods across all four corruptions.}
    \vspace{-0.2cm}
    \begin{adjustbox}{max width=\linewidth}
    \begin{tabular}{r|r| c|cccc|c}
    \toprule
    \textbf{Initial} & \textbf{Backbone} & 
    \textbf{mCE} & 

   \textbf{Fog} & 
   \textbf{Wet} &
   \textbf{Snow} & 
   \textbf{Beam} &
   \textbf{Avg}
    \\
    \midrule\midrule
    \rowcolor{yellow!10}
    \textit{sup.-only}  & MinkUNet & 100.00&56.11 & 55.29 &52.04 & 57.19& 55.16\\

    R2DM~\cite{nakashima2024r2dm}  &MinkUNet & 92.59 & 58.23& 60.00&54.42 & 61.18&58.46 \\
    Text2LiDAR~\cite{wu2024text2lidar}  &MinkUNet & 96.17 &55.65 & 60.35& 51.72&59.64 &56.84   \\
    WeatherGen~\cite{wu2025weathergen}  &MinkUNet &93.11 &56.97 &60.00 & 54.68 & 61.29 &58.23 \\
    \rowcolor{omnilidar}
    \textbf{OmniLiDAR}   &MinkUNet & \textbf{91.02} & \textbf{59.14} & \textbf{61.34} & \textbf{54.79} & \textbf{61.37} & \textbf{59.16} \\
    \bottomrule
    \end{tabular}
    \end{adjustbox}
\label{tab:robustness}
\vspace{-0.2cm}
\end{table}
\begin{table*}[t]
    \centering
    \caption{Generative data augmentation (GDA) results for 3D object detection on the Quadruped platform of \textit{Pi3DET} dataset, evaluated using R11@0.5.}
    \vspace{-0.2cm}
    \begin{adjustbox}{max width=\textwidth}
    \begin{tabular}{r|r|cc|cc|cc|cc|cc|cc}
    \toprule
    \multirow{2}{*}{\textbf{Method}} & \multirow{2}{*}{\textbf{Backbone}} & \multicolumn{2}{c|}{\textbf{1\%}} & \multicolumn{2}{c|}{\textbf{5\%}} &
    \multicolumn{2}{c|}{\textbf{10\%}} &
    \multicolumn{2}{c|}{\textbf{20\%}} &
    \multicolumn{2}{c|}{\textbf{50\%}} &
    \multicolumn{2}{c}{\textbf{100\%}}
    \\
    & & Vehicle  & Pedestrian  &  Vehicle  & Pedestrian  &Vehicle  & Pedestrian  & Vehicle  & Pedestrian  & Vehicle  & Pedestrian  & Vehicle  & Pedestrian   
    \\
    \midrule\midrule

    \textit{sup.-only} & PVRCNN & 0.03 & 0.07 &16.48 & 36.70 & 21.66 & 35.89 & 23.07 & 33.47 & 27.10 & 36.05 & 27.68 & 39.67
    \\
    \rowcolor{omnilidar}
    \textbf{OmniLiDAR} & PVRCNN  & \textbf{8.49} & \textbf{24.00} & \textbf{26.50} & \textbf{38.37} & \textbf{30.84} & \textbf{40.68} & \textbf{34.49} & \textbf{42.89} & \textbf{35.24} & \textbf{41.69} & \textbf{32.46} & \textbf{40.74}
    \\

    \textit{sup.-only} & VoxelRCNN & 0.04 & 0.91 &21.09 & 35.90 & 26.20 & 38.92 & 28.13 & 39.93 & 32.78 & 38.78 & 30.59 & 41.44 
    \\
    \rowcolor{omnilidar}
    \textbf{OmniLiDAR} & VoxelRCNN  & \textbf{20.46} & \textbf{33.05} & \textbf{30.40} & \textbf{40.50} & \textbf{35.99} & \textbf{42.35} & \textbf{36.84} & \textbf{42.04} & \textbf{38.24} & \textbf{42.98} & \textbf{38.35} & \textbf{42.31} 
    \\
    \bottomrule
\end{tabular}
\end{adjustbox}
\label{tab:quadruped_det}
\vspace{-0.3cm}
\end{table*}
\begin{table}[t]
    \centering
    \caption{Generative data augmentation (GDA) results for 3D object detection on the Drone platform of \textit{Pi3DET} dataset, evaluated using R11@0.5. Veh. denotes the Vehicle class.}
    \vspace{-0.2cm}
    \begin{adjustbox}{max width=0.48\textwidth}
    \begin{tabular}{r|r|c|c|c|c|c|c}
    \toprule
    \multirow{2}{*}{\textbf{Method}} & \multirow{2}{*}{\textbf{Backbone}} & \textbf{1\%} & \textbf{5\%} &
    \textbf{10\%} &
    \textbf{20\%} &
    \textbf{50\%} &
    \textbf{100\%}
    \\
    & & Veh.  &  Veh.    &Veh.   & Veh.   & Veh.   & Veh.     
    \\
    \midrule\midrule
    \textit{sup.-only} & PVRCNN & 9.09  & 20.06  & 21.39  & 25.64  & 28.11 &  29.05 
    \\
    \rowcolor{omnilidar}
    \textbf{OmniLiDAR} & PVRCNN  & \textbf{19.66} & \textbf{44.70} & \textbf{46.30} & \textbf{46.73} & \textbf{49.25} & \textbf{49.79}
    \\
    \textit{sup.-only} & VoxelRCNN & 0.19  & 23.27  & 25.15  & 32.01 &  35.30 &  35.52 
    \\
    \rowcolor{omnilidar}
    \textbf{OmniLiDAR} & VoxelRCNN  & \textbf{29.65} & \textbf{47.87} & \textbf{49.99} & \textbf{50.34} & \textbf{49.94} & \textbf{50.20}
    \\
    \bottomrule
\end{tabular}
\end{adjustbox}
\label{tab:drone_det}
\vspace{-0.3cm}
\end{table}
\begin{table}[!t]
    \centering
    \caption{Effect of reduced beam configuration on LiDAR semantic segmentation on the \textit{nuScenes} dataset under the Beam-32 setting. Results with supervised-only training and generative data augmentation are reported in mIoU (\%).}
    \vspace{-0.2cm}
    \resizebox{0.94\linewidth}{!}{
    \begin{tabular}{r|r|cccc}
        \toprule
        \textbf{Method} & \textbf{Backbone}& $\mathbf{1\%}$ & $\mathbf{10\%}$ & $\mathbf{20\%}$ & $\mathbf{100\%}$
        \\\midrule\midrule
  
         \textit{Sup.-only} & SPVCNN &  $30.45$ &  $60.23$ &  $67.26$ & $75.34$ \\
         \rowcolor{omnilidar}
         \textbf{OmniLiDAR} & SPVCNN & \textbf{38.38} &  \textbf{62.95} & \textbf{68.30} & \textbf{75.52} \\

         \textit{Sup.-only} &  MinkUNet & 33.27 & 58.94 & 67.15 & 75.02 
        \\
        \rowcolor{omnilidar} 
        \textbf{OmniLiDAR} & MinkUNet & \textbf{39.68} & \textbf{64.11} & \textbf{69.62} & \textbf{75.25}
        \\\bottomrule
    \end{tabular}}
     \label{tab:downstream_seg_nusc}
    \vspace{-0.2cm}
\end{table}

\begin{table}[t]
    \centering
    \caption{Inference efficiency comparison of LiDAR generation methods on the \textit{KITTI-360} dataset. We report the average inference time per frame in seconds (I.T.(s)) measured on an H100 GPU, along with model size and FRD for reference.}
    \vspace{-0.2cm}
    \begin{adjustbox}{max width=0.98\linewidth}
    \begin{tabular}{r|c|c|c|c}
    \toprule
    \textbf{Method} & 
    \textbf{Representation} & 
    \textbf{FRD$\downarrow$}  &
   \textbf{Params.(M)} & 
   \textbf{I.T.(s)} 
    \\
    \midrule\midrule
    LiDARGen~\cite{nakashima2024r2dm} & Range Image & 579.39 & \textbf{29.69} & 18.36
    \\
    R2DM~\cite{nakashima2024r2dm} & Range Image & 179.43 & 31.10 & 2.56
    \\
    Text2LiDAR~\cite{wu2024text2lidar} & Range Image & 425.90 & 45.77 & 4.90
    \\
    WeatherGen~\cite{wu2025weathergen} & Range Image & 160.20 & 31.71 & \textbf{1.87}
    \\
    \rowcolor{omnilidar}
    \textbf{OmniLiDAR} & Range Image & \textbf{158.13} & 97.73 & 3.52
    \\
    \bottomrule
    \end{tabular}
    \end{adjustbox}
\label{tab:latency}
\vspace{-0.4cm}
\end{table}

\subsection{Comparative Study}
\label{sec:comparative}
\noindent\textbf{Comparison with State-of-the-Art Methods.}
We compare OmniLiDAR with recent diffusion-based LiDAR generation methods on a unified multi-domain evaluation suite. Following standard practice, baseline methods are trained as separate models for each domain, whereas OmniLiDAR is trained once as a single unified model across all domains. As summarized in \Cref{tab:comparative_domains}, OmniLiDAR achieves consistently competitive generation performance across all evaluated domains, demonstrating that a single diffusion model can effectively model heterogeneous LiDAR distributions induced by diverse sensing conditions.

We observe relatively higher metric values on the Drone and Quadruped domains, which can be attributed to the domain mismatch between vehicle-centric LiDAR data used to train the feature-based metric extractors and the sensing geometries of aerial and quadruped platforms. Despite this bias, the metrics remain suitable for relative comparison across methods within each domain. Meanwhile, MMD and JSD are computed from occupancy distribution statistics (\emph{e.g.}, histograms) without relying on learned feature extractors, and OmniLiDAR achieves competitive MMD/JSD scores on all domains. In addition, we report results on the widely used KITTI-360 benchmark in \Cref{tab:kitti360_gen}. Under this conventional single-dataset evaluation setting, OmniLiDAR remains competitive with existing methods, despite being trained as a unified multi-domain model.

\noindent\textbf{Semantic Segmentation with Generative Data Augmentation.}
LiDAR semantic segmentation requires dense point-level annotations, which are costly to acquire at scale. As a result, limited-label regimes are common in practice. We evaluate the effectiveness of generative data augmentation for LiDAR semantic segmentation on SemanticKITTI under varying levels of supervision. As reported in \Cref{tab:semkitti_seg}, generative augmentation does not universally improve segmentation performance. Its benefit strongly depends on the fidelity of the generated samples. In particular, existing LiDAR generation methods such as LiDARGen yield limited or inconsistent gains over supervised-only training.

In contrast, OmniLiDAR consistently improves mIoU across all evaluated data regimes (1\%, 10\%, and 20\%). These improvements hold across both voxel-based backbones (MinkUNet~\cite{choy2019minkunet}) and voxel--point fusion backbones (SPVCNN~\cite{tang2020spvcnn}). This observation suggests that the gains are not tied to a specific segmentation architecture. Instead, the results indicate that high-quality, semantically faithful, and geometrically consistent LiDAR generation is critical for effective data augmentation in low-label settings. We further observe a consistent gain on the real-world adverse-weather SemanticSTF benchmark (\Cref{tab:semanticstf_lidarseg}), supporting the effectiveness of OmniLiDAR-generated weather samples beyond synthetic corruption settings.

\noindent\textbf{Robustness Evaluation under Adverse Conditions.}
We evaluate whether generative data augmentation improves robustness when test-time conditions deviate from the training distribution using the SemanticKITTI-C benchmark. \Cref{tab:robustness} reports results under multiple out-of-distribution corruptions, including fog, snow, wet ground, and beam missing. Compared to supervised-only training, augmenting the training set with generated LiDAR samples consistently improves robustness, resulting in lower mCE and higher mIoU across corruption types. OmniLiDAR achieves the best overall robustness, with the lowest mCE (91.02) and the highest average mIoU (59.16). These gains are consistently observed across the evaluated corruption types, indicating that OmniLiDAR provides effective robustness-oriented augmentation under adverse conditions.

\noindent\textbf{3D Object Detection with Generative Data Augmentation.}
LiDAR-based object detection on non-vehicle platforms, such as quadruped robots and aerial drones, is often constrained by limited training data due to the cost and complexity of data collection and annotation. We evaluate whether generative data augmentation can alleviate this limitation by assessing 3D object detection performance on the Quadruped and Drone benchmarks of Pi3DET. \Cref{tab:quadruped_det} and \Cref{tab:drone_det} compare supervised-only training with generative augmentation using OmniLiDAR-generated samples.

On the Quadruped benchmark, incorporating OmniLiDAR-generated samples consistently improves detection performance over supervised-only training across all labeled data regimes, from 1\% to 100\%. These improvements are observed with both PVRCNN~\cite{shi2020pv} and VoxelRCNN~\cite{deng2021voxel}, indicating that the gains are not specific to a particular detection architecture. Similar trends are observed on the Drone benchmark, where generative augmentation with OmniLiDAR also leads to improved detection accuracy under the same evaluation setting.

\begin{figure}[t]
    \centering
    \includegraphics[width=0.48\textwidth]{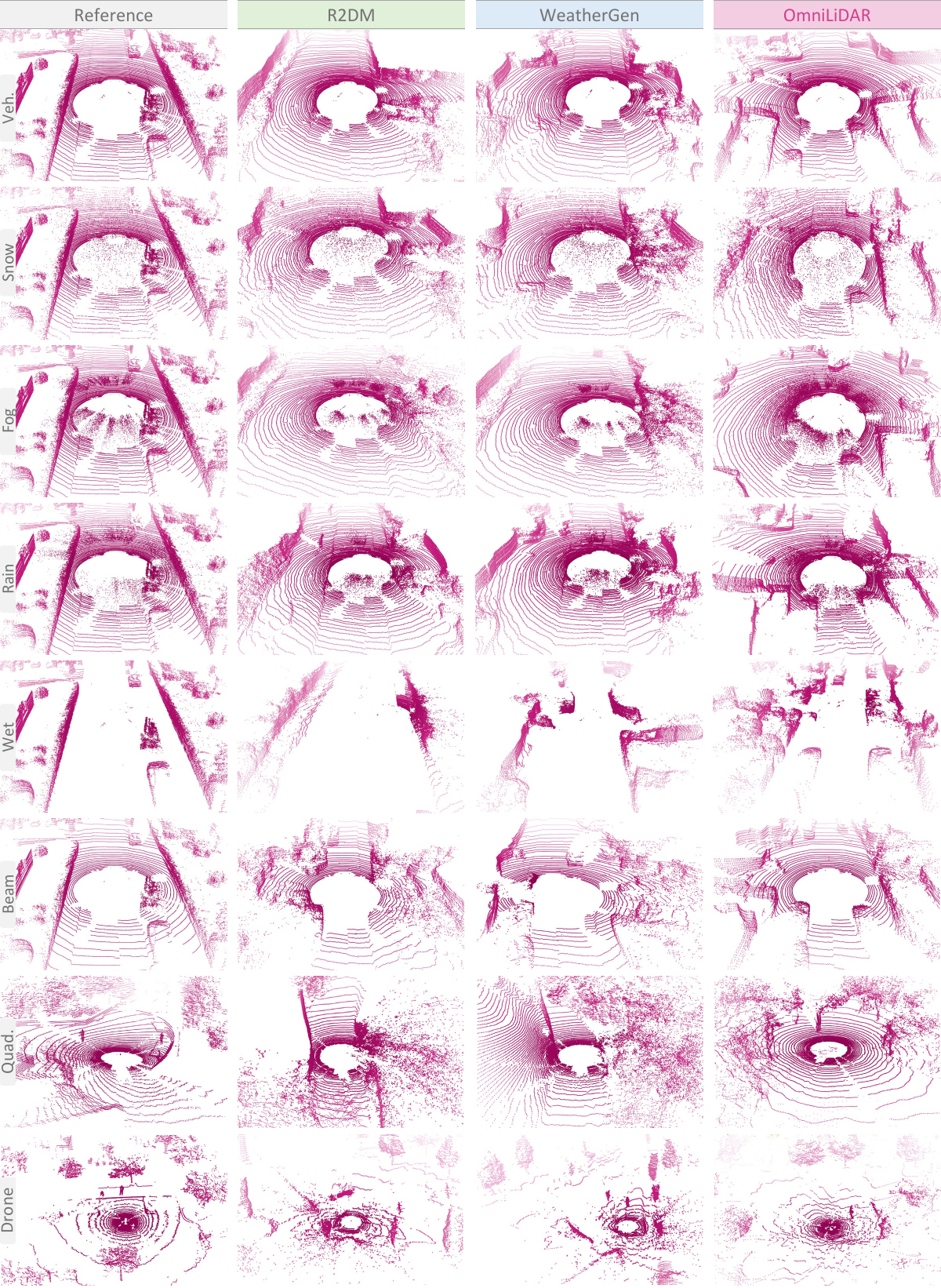}
    \vspace{-0.2cm}
    \caption{Qualitative comparisons of LiDAR scene generation across eight domains. Each row corresponds to a different domain (Vehicle, Snow, Fog, Rain, Wet Ground, Beam-32, Quadruped, and Drone), and each column shows results from different methods. From left to right: reference LiDAR scans, R2DM, WeatherGen, and OmniLiDAR.}
    \label{fig:comprative_scene}
\end{figure}

\noindent\textbf{Effect of Reduced Beam Configuration.}
We evaluate generative data augmentation under reduced-beam sensing by assessing semantic segmentation performance in a 32-beam setting. Experiments are conducted on nuScenes~\cite{caesar2020nuscenes}, which is captured with a 32-beam LiDAR, making it a natural testbed for this analysis. \Cref{tab:downstream_seg_nusc} compares supervised-only training with generative augmentation using Beam-32 samples generated by OmniLiDAR. Across both SPVCNN and MinkUNet backbones, incorporating generated samples consistently improves segmentation performance over supervised-only training. These results indicate that samples generated by OmniLiDAR provide effective complementary supervision under reduced beam configurations.

\begin{figure}[t]
    \centering
   \includegraphics[width=0.42\textwidth]{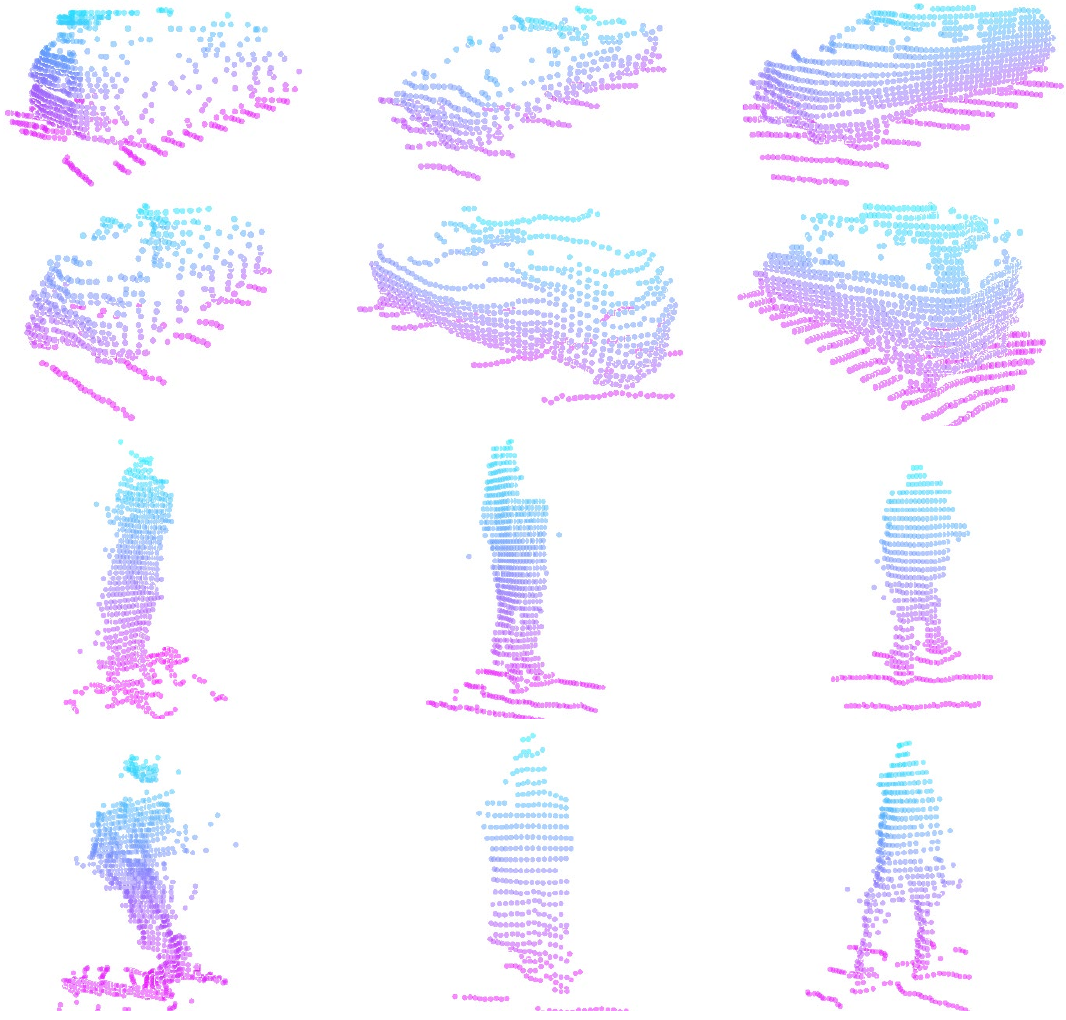}
    \caption{Qualitative comparison of object-level LiDAR geometry.
    Objects extracted from generated scenes are shown from left to right: R2DM~\cite{nakashima2024r2dm}, WeatherGen~\cite{wu2025weathergen}, and OmniLiDAR.}
    \label{fig:comprative_objects}
\end{figure}

\begin{figure}[t]
    \centering
   \includegraphics[width=\linewidth]{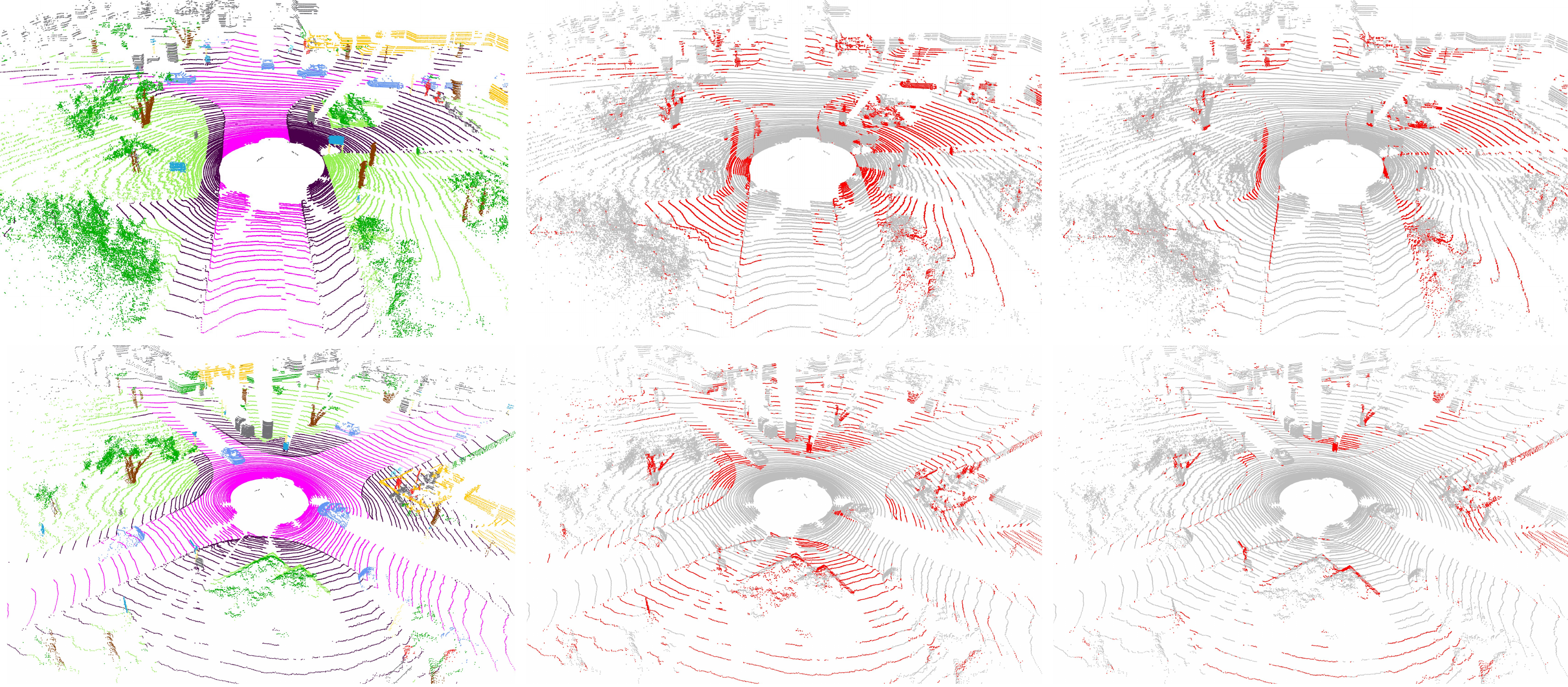}
    \vspace{-0.3cm}
    \caption{Qualitative comparison of LiDAR semantic segmentation on SemanticKITTI under 1\% labeled data using SPVCNN. From left to right: ground-truth annotations, supervised-only predictions, and predictions trained with GDA using OmniLiDAR-generated samples. \textcolor{gray}{Correct} / \textcolor{red}{incorrect} predictions are highlighted in \textcolor{gray}{gray} / \textcolor{red}{red}, respectively.}
    \label{fig:comprative_kiit}
\end{figure}

\begin{figure}[t]
    \centering
   \includegraphics[width=0.48\textwidth]{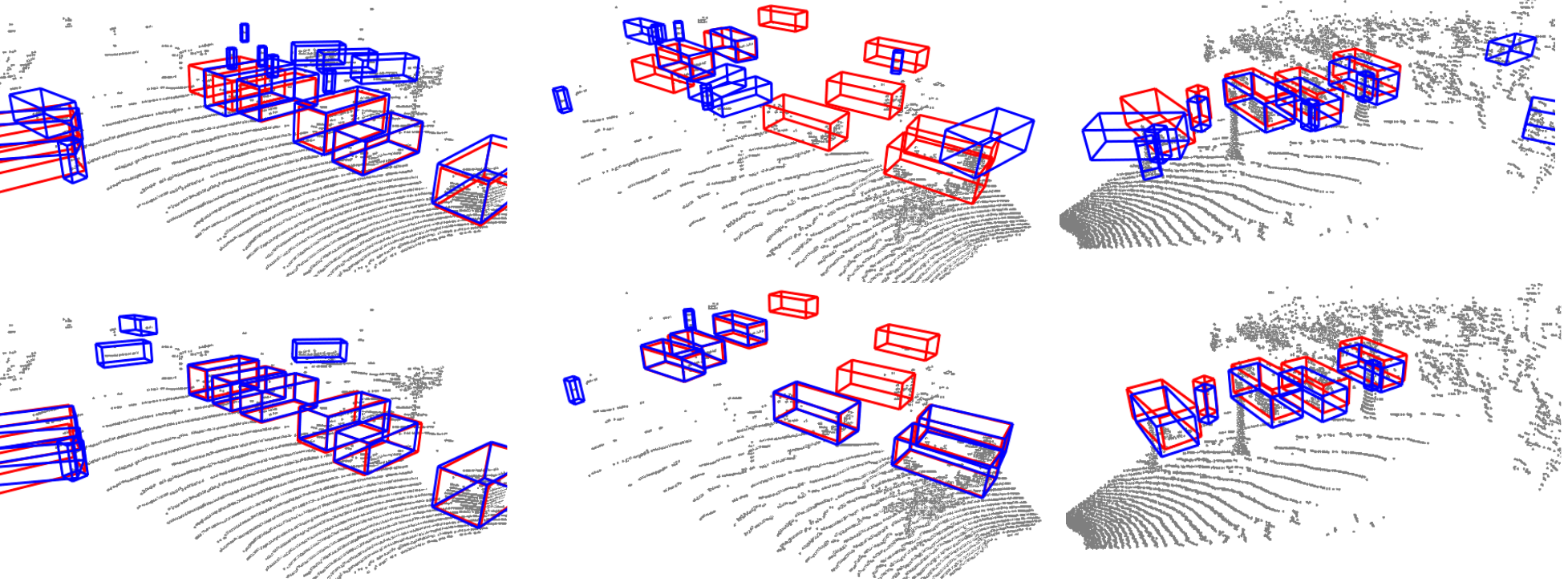}
    \vspace{-0.3cm}
    \caption{Qualitative 3D detection results on Pi3DET~\cite{liang2025perspective} with VoxelRCNN under 10\% labeled data. Top: supervised-only training.
    Bottom: training augmented with synthetic LiDAR samples generated by OmniLiDAR.
    \textcolor{blue}{Blue} and \textcolor{red}{red} boxes denote  \textcolor{blue}{predictions} and \textcolor{red}{ground truth}, respectively.}
    \label{fig:comprative_det}
\end{figure}

\begin{table*}[t]
    \centering
    \caption{Ablation study of OmniLiDAR evaluating different training configurations and components on LiDAR generation quality across domains. Distribution-based metrics (FRD, FRID, FPVD, FSVD, JSD, and MMD) are reported for different weather, sensor, and platform settings. Lower values indicate better performance. The MMD metric is reported in $10^{-4}$.}
    \vspace{-0.2cm}
    \begin{adjustbox}{max width=\textwidth}
    \begin{tabular}{
        r|r|
        cccccc|cccccc|cccccc|cccccc
    }
    \toprule
    \multirow{2}{*}{\textbf{ID}} 
    & \multirow{2}{*}{\textbf{Configuration}} 
    & \multicolumn{6}{c|}{\textbf{Vehicle}} 
    & \multicolumn{6}{c|}{\textbf{Snow}} 
    & \multicolumn{6}{c|}{\textbf{Fog}}
    & \multicolumn{6}{c}{\textbf{Rain}} 
    \\
    & & FRD & FRID & FPVD & FSVD & JSD & MMD
      & FRD & FRID & FPVD & FSVD & JSD & MMD
      & FRD & FRID & FPVD & FSVD & JSD & MMD
      & FRD & FRID & FPVD & FSVD & JSD & MMD
    \\
    \midrule\midrule

    \rowcolor{yellow!10}
     1 & Single-Domain  & 443.31&10.69 &11.27 &12.75 &0.04 & 2.97 & 453.14 &11.25 & 28.76 & 27.02 & 0.05 &3.32 &444.22 &11.56 &20.67 &17.17 &0.04 &4.28 &418.60 &8.52 &14.75 &15.61 & \textbf{0.03} &1.12 \\

    2 &Batch-Homogeneous &
     430.89 & 9.36 & 9.95 & 11.41 & 0.04 & \textbf{0.47} & 434.13 & 7.68 & 21.21 & 18.99 & \textbf{0.04} & 0.91 & 431.63 & 9.52 & 18.01 & 15.99 & \textbf{0.03} & 0.70 & 432.34 & 6.55 & 15.27 & 15.21 & 0.04 & 1.05\\

    3& CDTS & 420.50 & 8.87 & 8.84 & 11.33 & \textbf{0.03} & 0.61 & 428.88 & 5.09 & 20.72 & 19.21 & \textbf{0.04} & \textbf{0.69} & 427.36 & 8.70 & 17.51 & 15.10 & 0.04 & 1.02 & 417.13 & 6.40 & 13.99 & 14.33 & 0.04 & 1.51\\

    4& CDTS + CDFM &419.46 & 8.39 & 8.71 & 11.00 & \textbf{0.03} & 0.57 &442.86 & 6.65 & 19.25 & 17.69 & 0.05 & 1.11 &419.44 & 8.34 & 18.28 & 16.04 & \textbf{0.03} & 0.78 &
    411.61 & 5.51 & 14.33 & 14.59 & \textbf{0.03} & \textbf{0.56} \\

    5& CDTS + DAFS & 416.63 & 8.83 & 9.46 & 11.72 & \textbf{0.03} & 0.90 &426.44 & \textbf{4.46} & 19.44 & 18.42 & \textbf{0.04} & 0.89 &418.57 & 7.78 & 15.17 & 13.75 & \textbf{0.03} & 0.75 &
    420.20 & 5.76 & 14.67 & 14.57 & \textbf{0.03} & 0.94 \\

     \rowcolor{omnilidar}
     6 &  \textbf{Full Model} & \textbf{410.91} & \textbf{8.37} & \textbf{7.79} & \textbf{10.06} & 0.04 & 0.99 & \textbf{424.02} & 4.76 & \textbf{15.68} & \textbf{14.94} & 0.05& 0.86 & \textbf{396.64} & \textbf{7.21} & \textbf{15.10} & \textbf{13.40} & \textbf{0.03} & \textbf{0.54} & \textbf{402.73} & \textbf{5.42} & \textbf{12.02} & \textbf{12.62} &0.04 & 0.98\\
    
    \midrule\midrule
    
    \multirow{2}{*}{\textbf{ID}} 
    & \multirow{2}{*}{\textbf{Configuration}} 
    & \multicolumn{6}{c|}{\textbf{Wet Ground}} 
      & \multicolumn{6}{c|}{\textbf{Beam-32}} 
      & \multicolumn{6}{c|}{\textbf{Drone}}
      & \multicolumn{6}{c}{\textbf{Quadruped}} 
    \\
    & & FRD & FRID & FPVD & FSVD & JSD & MMD
      & FRD & FRID & FPVD & FSVD & JSD & MMD
      & FRD & FRID & FPVD & FSVD & JSD & MMD
      & FRD & FRID & FPVD & FSVD & JSD & MMD
    \\ \midrule\midrule
    
     \rowcolor{yellow!10}
     1 & Single-Domain  & 455.14 &6.07 & 13.42& 16.47& \textbf{0.04} &\textbf{0.94} &396.99 &5.69 &13.27 &13.19 & \textbf{0.03} &1.00 & 1795.78 & 88.82 & 37.01 & 29.77& 0.07&1.98 &425.50 & 5.81 & 30.42 &26.78 &0.08 &1.86 \\

     2 & Batch-Homogeneous & 458.49 & 5.33 & 12.12 & 15.23 & 0.05 & 2.07 & 383.32 & 5.72 & 12.79 & 11.32 & 0.04 & 0.69 &
    1788.85 & 83.06 & 36.30 & 31.30 & \textbf{0.05} & 1.03 & 440.93 & 4.58 & 29.84 & 23.07 & \textbf{0.06} & 0.78 \\

    3& CDTS & \textbf{443.44} & \textbf{4.62} & 11.40 & 14.86 & 0.05 & 1.27 &374.94 & 5.30 & 11.13 & 10.61 & 0.04 & 0.76 &1767.75 & 80.71 & 33.71 & 31.35 & \textbf{0.05} & \textbf{1.02} &429.86 & 3.56 & 29.47 & 25.20 & \textbf{0.06} & 1.20 \\

    4& CDTS + CDFM & 451.96 & 4.74 & 11.31 & 14.81 & \textbf{0.04} & 1.34 &365.71 & 4.75 & 10.78 & 10.64 & \textbf{0.03} & \textbf{0.51} &1790.69 & 80.62 & \textbf{23.15} & \textbf{22.42} & 0.06 & 1.10 &
    \textbf{419.64} & \textbf{3.37} & 27.62 & 24.01 & \textbf{0.06} & \textbf{0.62} \\
    
    5& CDTS + DAFS & 448.60 & 4.93 & 11.87 & 14.82 & 0.05 & 1.27 &372.87 & 5.41 & 11.75 & 11.25 & \textbf{0.03} & 0.94 &
    1780.49 & \textbf{79.79} & 33.83 & 27.41 & 0.06 & 1.24 & 438.19 & 5.26 & 31.55 & 26.96 & 0.07 & 1.10 \\

     \rowcolor{omnilidar}
     6 & \textbf{Full Model} & 448.83 &4.73 & \textbf{9.87} & \textbf{12.40} &0.05 &1.21 & \textbf{352.61} & \textbf{3.98} & \textbf{9.21} & \textbf{8.99} & 0.04& 0.90 & \textbf{1748.28} & 81.09 & 31.81& 26.34& 0.06& 1.60& 422.44&4.38 &\textbf{27.59} & \textbf{23.01} &0.07 &0.87 \\
    
    \bottomrule
    \end{tabular}
    \end{adjustbox}
    \label{tab:ablation}
\end{table*}
\noindent\textbf{Inference Efficiency.}
We analyze the inference efficiency of OmniLiDAR to assess its practical deployment cost. \Cref{tab:latency} reports the average per-frame inference time of different LiDAR generation methods on the KITTI-360 dataset, measured on an NVIDIA H100 GPU. Despite having a larger model size, OmniLiDAR achieves inference latency comparable to existing methods, indicating that increased model capacity does not necessarily incur prohibitive computational overhead at inference time. These results suggest that OmniLiDAR maintains practical inference efficiency while delivering improved generation quality.

\noindent\textbf{Qualitative Analysis.}
We present qualitative results to complement the quantitative evaluation and to illustrate how OmniLiDAR models heterogeneous LiDAR domains. \Cref{fig:comprative_scene} compares generated scans across eight representative domains, spanning adverse weather, sensor-configuration shifts (Beam-32), and cross-platform acquisition (drone and quadruped), against separately trained state-of-the-art LiDAR generation methods. OmniLiDAR generates scans with coherent scene layout and domain-consistent sampling patterns. Under adverse weather, the generated returns exhibit increased sparsity and structured missing-return patterns while preserving salient geometric structures.
Under Beam-32 and cross-platform settings, OmniLiDAR more closely reflects the target sampling patterns and acquisition geometries, whereas separately trained methods can exhibit artifacts or viewpoint/occlusion patterns that deviate from the intended sensing configuration. In addition, \Cref{fig:comprative_objects} indicates higher object-level fidelity, with more complete object contours and more plausible local surface structures than those produced by separately trained methods.

We further visualize downstream effects in \Cref{fig:comprative_kiit,fig:comprative_det}. In \Cref{fig:comprative_kiit}, LiDAR segmentation models trained with data augmentation using samples generated by OmniLiDAR produce more spatially coherent predictions, with fewer spurious regions, than supervised-only training under limited supervision. In \Cref{fig:comprative_det}, generative augmentation using OmniLiDAR improves detection results on cross-platform benchmarks, with higher recall and more stable localization under platform shifts.

\subsection{Ablation Study}
\label{sec:ablation}
We analyze the contribution of different training strategies and architectural components in OmniLiDAR by following the progressive design of our unified diffusion framework. We begin with \textit{single-domain} training, where an independent model is trained separately for each domain, resulting in eight domain-specific generators without cross-domain parameter sharing. We then consider a \textit{unified} model trained once across all domains with \textit{batch-homogeneous} sampling, where each mini-batch contains samples from a single domain. With a shared range-image representation, this unified formulation improves generation quality over single-domain training across most domains (\emph{e.g.}, Vehicle FRD: 443.31$\rightarrow$430.89; Fog MMD: 4.28$\rightarrow$0.70), demonstrating that parameter sharing under a common representation is beneficial. Nevertheless, batch-homogeneous sampling enforces domain isolation at the mini-batch level, which limits direct cross-domain interaction during optimization.

To overcome this limitation, we introduce the \textbf{Cross-Domain Training Strategy (CDTS)}, which removes the batch-homogeneous constraint and enables samples from different domains to be jointly optimized within each mini-batch. As shown in \Cref{tab:ablation}, CDTS yields consistent gains across multiple domains and metrics (\emph{e.g.}, Beam-32 FRD: 383.32$\rightarrow$374.94; Drone FRID: 83.06$\rightarrow$80.71). Given its consistent improvements over batch-homogeneous training, CDTS is adopted as the base configuration for subsequent ablations.

We progressively incorporate \textbf{Cross-Domain Feature Modeling (CDFM)} and \textbf{Domain-Adaptive Feature Scaling (DAFS)} to address complementary aspects of multi-domain generation. CDFM is designed to model geometry-aligned long-range dependencies shared across domains, leading to improved structural coherence of generated scans. In \Cref{tab:ablation}, its effect is particularly evident on geometry-sensitive settings such as Beam-32, Drone, and Quadruped, where preserving scan-aligned long-range structure is especially important. This is reflected by improvements in multiple geometric metrics, particularly FRD/FPVD on Beam-32, FPVD/FSVD on Drone, and FRD/FPVD/FSVD on Quadruped. In contrast, DAFS targets domain-specific feature statistics arising from heterogeneous conditions (\emph{e.g.}, adverse weather and sensor-related shifts) and contributes to more stable generation under such settings (\emph{e.g.}, Fog FSVD: 16.04$\rightarrow$13.75). Combining all components results in the full model, which achieves the best overall performance across most domains, validating the effectiveness of the proposed design.

\section{Conclusion}
\label{sec:conclusion}

In this paper, we present OmniLiDAR, a unified text-prompt-conditioned diffusion framework for controllable LiDAR scene generation, enabling a single model to capture distribution shifts induced by weather, sensor configurations, and acquisition platforms. To enable stable unified training, we introduce a cross-domain training strategy together with geometry-aligned feature modeling and domain-adaptive feature scaling, which jointly facilitate cross-domain parameter sharing while preserving domain-specific characteristics. We also construct an 8-domain LiDAR dataset under a unified protocol to support systematic evaluation. Extensive experiments show strong generation fidelity and consistent downstream benefits, particularly in limited-label and cross-domain settings. Overall, OmniLiDAR represents a practical step toward scalable and controllable LiDAR generative modeling for simulation and robust 3D perception.

\ifCLASSOPTIONcaptionsoff
  \newpage
\fi
\bibliographystyle{IEEEtran}
\bibliography{ref}

\end{document}